\documentclass[letterpaper, 10 pt, conference]{ieeeconf}
\IEEEoverridecommandlockouts

\usepackage[utf8]{inputenc}

\usepackage{mathbbol}

\usepackage{esvect}

\usepackage[percent]{overpic}

\usepackage{tabu}

\usepackage{multicol}

\usepackage{multirow}

\usepackage[overload]{empheq}

\usepackage{bm}

\usepackage{xcolor}
\definecolor{dg}{rgb}{0.1, 0.6, 0.2}       
\definecolor{b}{rgb}{0.0, 0.0, 1}          

\usepackage{epsfig}
\usepackage{float}
\usepackage{color}
\usepackage{url}

\usepackage{algorithmic}
\usepackage[ruled,linesnumbered,vlined]{algorithm2e}

\usepackage{amssymb, amsfonts}
\usepackage{amsmath}
\usepackage{mathrsfs}

\usepackage{amsthm}
\usepackage{mathtools}

\let\labelindent\relax
\usepackage{enumitem}       
\setenumerate[enumerate]{align=left}

\usepackage{graphicx}

\usepackage{cite}

\usepackage{stackengine}

\usepackage[flushleft]{threeparttable}

\usepackage{balance}

\newcommand{\norm}[1]{\left\lVert#1\right\rVert}

\makeatletter
\newlength\tmp@\newlength\t@mp
\newcommand{\comp}[3]
  {\mathop{ \settowidth\tmp@{$\displaystyle\mathop{#1}^{#3}_{#2}$}
  \hbox to \tmp@{\hss \settowidth\t@mp{$\displaystyle #1$}\setlength\t@mp{.45\t@mp}
  $\displaystyle\mathop{#1}^{\hspace\t@mp #3}_{\hspace{-\t@mp}#2}$
  \hss} }}
\makeatother

\newcommand{\Int}[2]
{\int_{#1}^{#2}}

\DeclareMathOperator*{\argmin}{argmin}

\usepackage[bookmarks=true]{hyperref}

\hypersetup{
    colorlinks=true,
    linkcolor=blue,
    filecolor=blue,      
    urlcolor=blue,
    citecolor=blue,
}

\def\R{\mathbb{R}}

\def\pos{p}

\newcommand{\vbf}[1]{{\bm{\mathbf{#1}}}}

\def\rot{R}
\def\tf{T}

\def\X{\mathcal{X}}
\def\Y{\mathcal{Y}}

\def\tf{\vbf{T}}
\def\rot{\vbf{R}}
\def\pos{\vbf{p}}

\usepackage[absolute,overlay]{textpos}
\usepackage{nccmath}    
\usepackage{multimedia} 

\begin{document}

\title{\bf ULOC: Learning to Localize in Complex Large-Scale Environments with Ultra-Wideband Ranges}

\author{
      Thien-Minh Nguyen$^{*1}$, \IEEEmembership{Member,~IEEE},
      Yizhuo Yang$^{*1}$,
      Tien-Dat Nguyen$^{2}$,
      \\
      Shenghai Yuan$^{1}$, \IEEEmembership{Member,~IEEE},
      Lihua Xie$^{1}$, \IEEEmembership{Fellow,~IEEE},
\thanks{$*$ Joint first authors}
\thanks{$^{1}$School of Electrical and Electronic Engineering, Nanyang Technological University, 50 Nanyang Avenue, Singapore 639798.}
\thanks{$^{1}$Faculty of Electrical and Electronic Engineering, Ho Chi Minh City University of Technology.}
\thanks{
This research is supported by the National Research Foundation, Singapore under its Medium Sized Center for Advanced Robotics Technology Innovation (CARTIN).
}
}

\maketitle

\thispagestyle{plain}
\pagestyle{plain}

\begin{abstract}
While UWB-based methods can achieve high localization accuracy in small-scale areas, their accuracy and reliability are significantly challenged in large-scale environments.
In this paper, we propose a learning-based framework named ULOC for Ultra-Wideband (UWB) based localization in such complex large-scale environments. First, anchors are deployed in the environment without knowledge of their actual position. Then, UWB observations are collected when the vehicle travels in the environment. At the same time, map-consistent pose estimates are developed from registering (onboard self-localization) data with the prior map to provide the training labels. We then propose a network based on MAMBA that learns the ranging patterns of UWBs over a complex large-scale environment. The experiment demonstrates that our solution can ensure high localization accuracy on a large scale compared to the state-of-the-art. We release our source code to benefit the community at \url{https://github.com/brytsknguyen/uloc}.
\end{abstract}

\IEEEpeerreviewmaketitle

\section{Introduction}

\begin{figure}
    \centering
    \includegraphics[width=\linewidth]{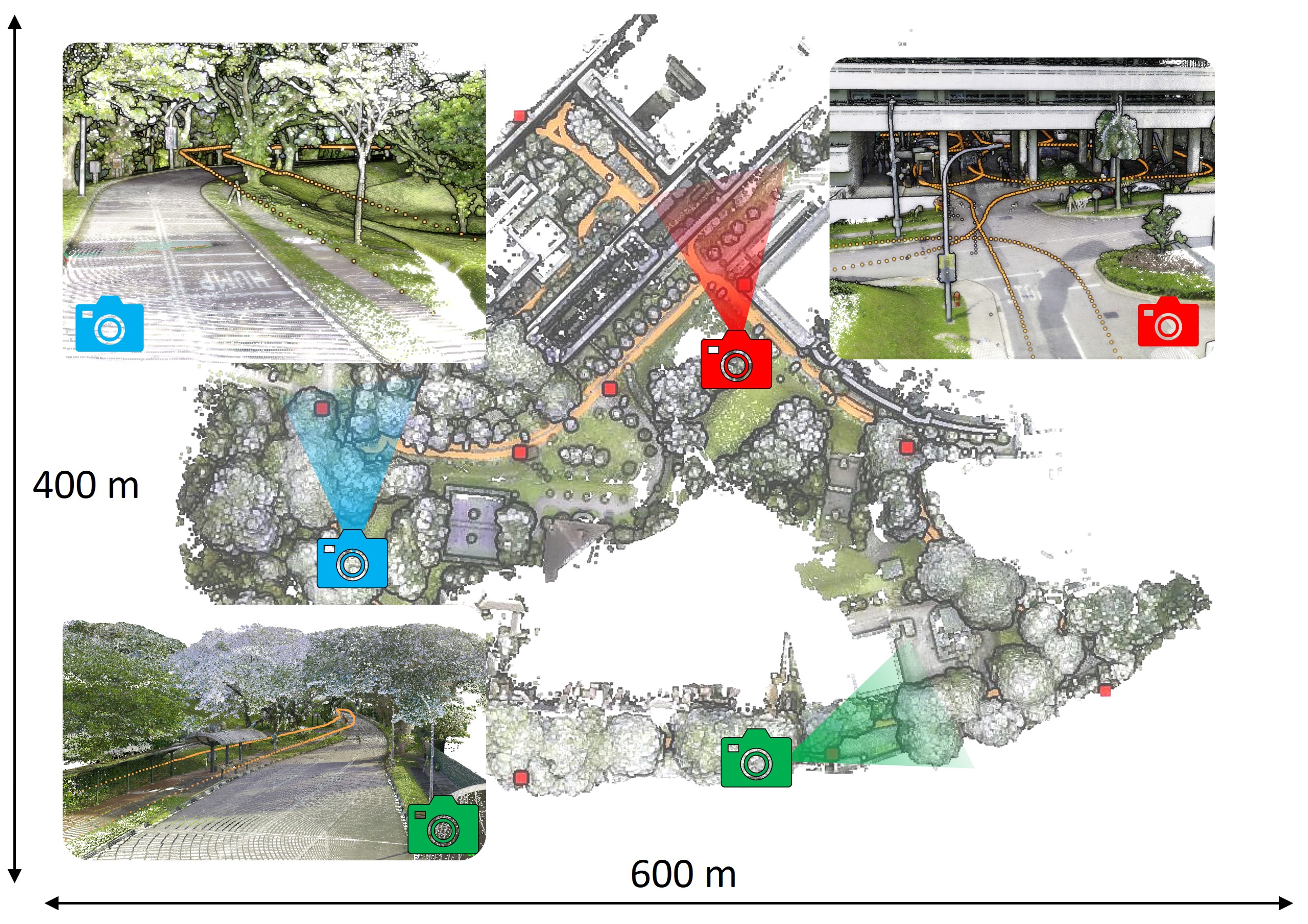}
    \includegraphics[width=\linewidth]{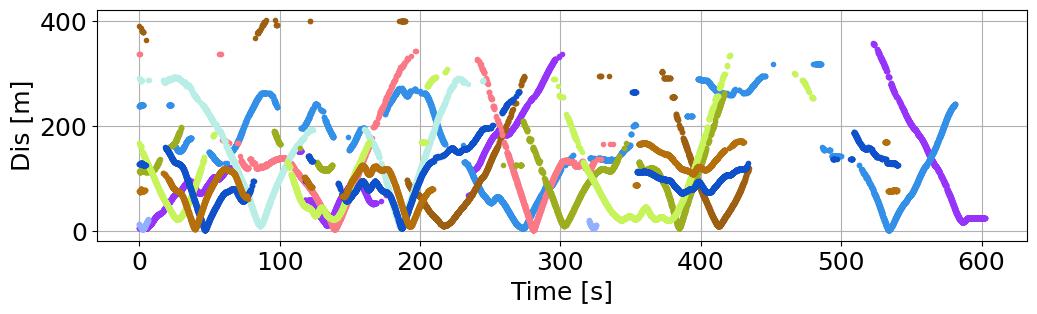}
    \includegraphics[width=\linewidth]{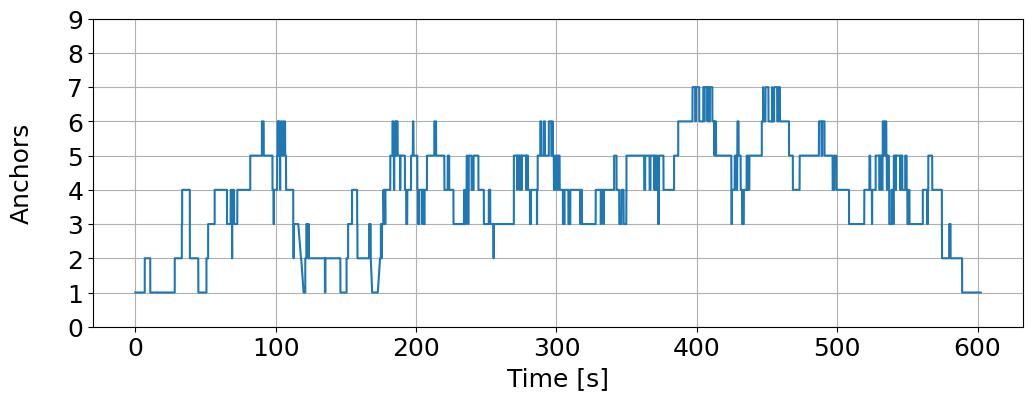}
\caption{Large scale environment with ten anchors (red squares) deployed over a 400m $\times$ 200m campus area. The path of the vehicle is marked in orange. On the bottom two plots, we show the UWB measurements to the anchors during the travel, and number of anchors seen over time on a 1s sliding window (bottom). We can see that  there is hardly any location where measurements to all ten anchors are available. Moreover, due to multi-path and NLOS conditions, the outliers and observation losses are also quite frequent. To the best of our knowledge, this environment being investigated in this paper is the largest with high-accuracy ground truth among existing works on UWB-based localization.}
\label{fig: large scale environment}
\vspace{-0.5cm}
\end{figure}

In recent years, Ultra-Wideband (UWB) has established its role as a promising solution for localization, finding applications in several areas such as logistics \cite{nguyen2023vr, jiang2023efficient}, swarming \cite{xu2022omni, zhou2022swarm, moron2022towards}, or traffic control \cite{wang2023stop}.
However, it is still challenging to expand it to applications in larger areas (Fig. \ref{fig: large scale environment}) such as seaports and airports.  

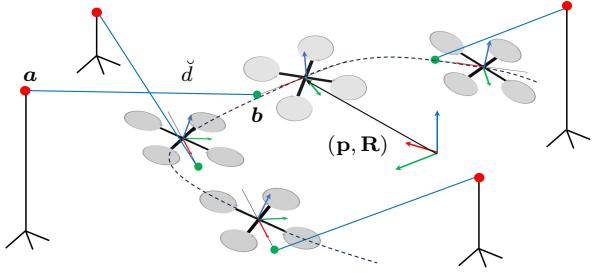
\begin{figure}
    \centering
    \begin{overpic}[width=0.9\linewidth,
                   ]{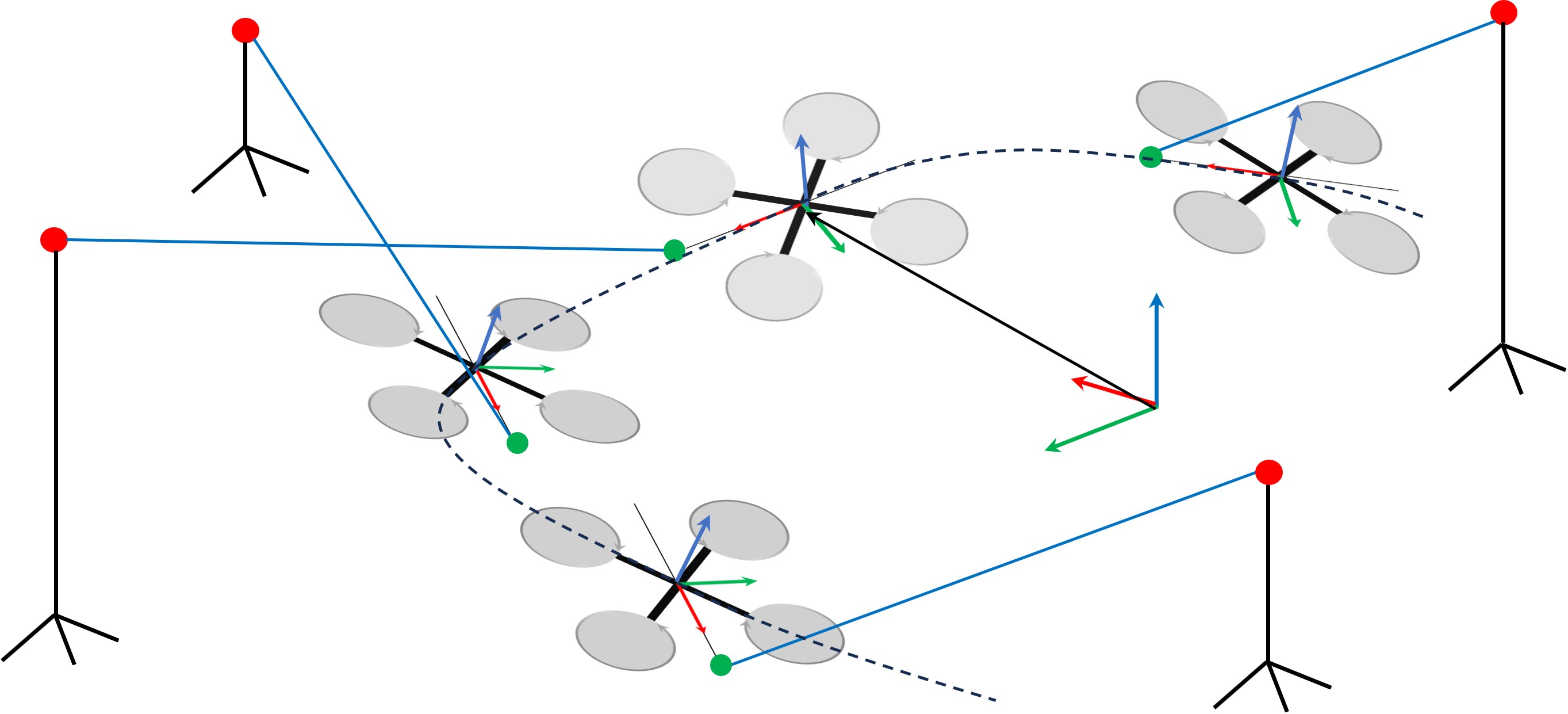}
                   \put(03, 32){\footnotesize $\bm{a}$}
                   \put(30, 32){\footnotesize $\breve{d}$}
                   \put(42, 25){\footnotesize $\bm{b}$}
                   \put(55, 20){\footnotesize $(\pos, \rot)$}
                   
    \end{overpic}
\caption{A simple UWB-based localization scheme in small-scale environments. If we define $\rot, \pos, d, \bm{a}, \bm{b}, a, b$ respectively as the robot orientation, its position, the range, anchor position, UWB tag's offset, and some scaling factors, the ranging model $d = a\norm{\pos + \rot\bm{b} - \bm{a}} + b$ can be used for full pose localization of UWB in small-scale environments. For the estimate to converge during a certain period of time, at least range measurements to each anchor have to be obtained.}
\label{fig: small scale environment}
\end{figure}

First, the anchor positions need to be obtained beforehand. 
Second, ranges to a minimum number of anchors over a certain period are required to ensure convergence of the position estimate.
While these issues can be resolved in small-scale environments (Fig. \ref{fig: small scale environment}), for example, the anchor positions can be obtained from motion capture system \cite{nguyen2016ultra}, or from total station \cite{nguyen2021ntuviral}. Since the area of operation is small, a sufficient number of anchors can also be installed to ensure reliable ranging to three anchors at all points. However, on a large scale, these measures quickly become less viable. Most importantly, issues such as non-line-of-sight (NLOS), multi-path, or unknown scaling factors also begin to manifest \cite{zhao2022util}, rendering UWB-based localization unattainable.

In recent years, onboard self-localization (OSL)-based solutions have offered an alternative solution to large-scale localization problems other than UWB. In many environments, a prior map can be acquired by the survey mapping technique or the offline OSL method itself. Since the map is mostly static, estimation errors can be bounded instead of accumulating in SLAM mode. Indeed, under this scheme the localization error is location-dependent, i.e., it only depends on the error at that location between the prior map and the real environment, as illustrated in Fig. \ref{fig: train vs test}.

While prior maps enable large-scale localization, achieving reliability in repetitive environments remains challenging for descriptor-based methods (DBM) \cite{yuan2023std, yin2024outram}. DBM requires high-dimensional data (images or point clouds) and is sensitive to environmental changes, while UWB localization faces issues like NLOS and interference. A promising solution is to combine OSL with UWB for effective complementary fusion.

This motivates us to consider UWB localization on a large scale via learning-based approaches. The rationales for this work are multifolds. First, UWB data is much more compact than other modalities, such as images and LiDAR. Second, the learning model can implicitly learn the anchor coordinates, scaling factors, and complex environmental effects such as multi-path, blockage, etc. Finally, we posit that not only the presence but also the absence of distance measurements to the anchors in the ranging sequence can contain learnable contexts about the location. Coupled with recent advances in natural language processing and time series learning models, which have demonstrated remarkable capabilities in exploiting such context, we propose a MAMBA-based \cite{gu2023mamba} learning model for this task, which demonstrates the ability to predict the position of the vehicle even when only one anchor is in LOS.



In summary, the contribution of this paper is as follows:

\begin{itemize}
    \item We propose a framework for UWB-based localization on large-scale complex environments leveraging OSL with prior maps.
    \item We propose a learning model based on MAMBA, which can effectively capture the contextual information from long UWB sequences for accurate localization. The designed network demonstrates superior performance compared to classical methods and a significant improvement over the existing state-of-the-art LSTM-based methods.
    \item The datasets, source code, and results are released for the benefit of the community.
\end{itemize}

The remainder of this paper is organized as follows: Sec. \ref{sec: related works} provide a review of related works. Sec. \ref{sec: methodology} goes into details of localization on the prior map and the design of our learning model. Sec. \ref{sec: experiments} presents the experimental results comparing ULOC with existing methods. Sec. \ref{sec: conclusion} concludes our work.

\section{Related Works} \label{sec: related works}

In the context of UWB-based localization, Mueller et al. \cite{mueller2015fusing} proposed a sensor fusion technique combining UWB with IMU data, enhancing accuracy but facing challenges in NLOS conditions. Nguyen et al. \cite{nguyen2016ultra} improved large-scale localization, but scalability issues arose with increasing nodes and robots. Tiemann et al. \cite{tiemann2017scalable} developed a scalable TDOA-based UWB system for multi-UAV navigation with high precision, though it was sensitive to anchor placement and environmental factors. Dai et al. \cite{dai2024cubature} later addressed NLOS errors with an improved Kalman Filter, enhancing accuracy but adding computational complexity. These classical approaches laid a strong foundation for UWB localization but are limited by scalability, environmental sensitivity, and challenges in ensuring accuracy and real-time performance.


Graph-based UWB localization methods provide enhances accuracy by optimizing sensor measurements in a graph structure, making them more robust in complex environments compared to classical methods. However, they come with several limitations. Wang et al. \cite{wang2017ultra} introduced a graph optimization framework for UWB localization, but it faced computational challenges in real-time scenarios. Fang et al. \cite{fang2019graph} improved this by integrating IMU data, enhancing accuracy but adding complexity due to the need for Lie-Manifold optimization. Zhou et al. \cite{zhou2021graph} applied a graph-optimization-based ZUPT/UWB fusion algorithm, achieving high precision by combining zero-velocity updates with UWB data, which enhanced robustness in dynamic environments. However, this method still requires careful calibration of parameters and sensor placement to avoid computational bottlenecks.  Overall, these methods offer better precision but struggle with high computational demands and implementation complexity.


Learning-based UWB localization methods leverage deep learning models to improve real-time deployment performance, particularly in challenging environments. While models like RoNet \cite{lim2019ronet} show high accuracy and real-time performance in specific environments, they struggle to generalize well to new settings or conditions without extensive retraining on large labeled datasets. DeepUWB \cite{nosrati2022improving} improves performance in cluttered environments but also faces generalization challenges, particularly when deployed in environments with different characteristics from the training data. This reliance on specific datasets limits the scalability and robustness of these methods in real-world, dynamic, and diverse scenarios.

Recent advancements in UWB localization focus on optimizing anchor placement \cite{zhao2022finding}, improving accuracy by strategic positioning. Many approaches also rely on UWB-IMU fusion to combine range and motion data, enhancing robustness in dynamic environments and NLOS conditions \cite{hashim2023uwb, hashim2023nonlinear}. Additionally, continuous representation methods model trajectories over time, yielding smoother and more accurate pose estimation \cite{li2023continuous, goudar2023continuous}, though they come with increased computational demands.



\section{Methodology} \label{sec: methodology}

\begin{figure}
    \centering
    \includegraphics[width=0.8\linewidth]{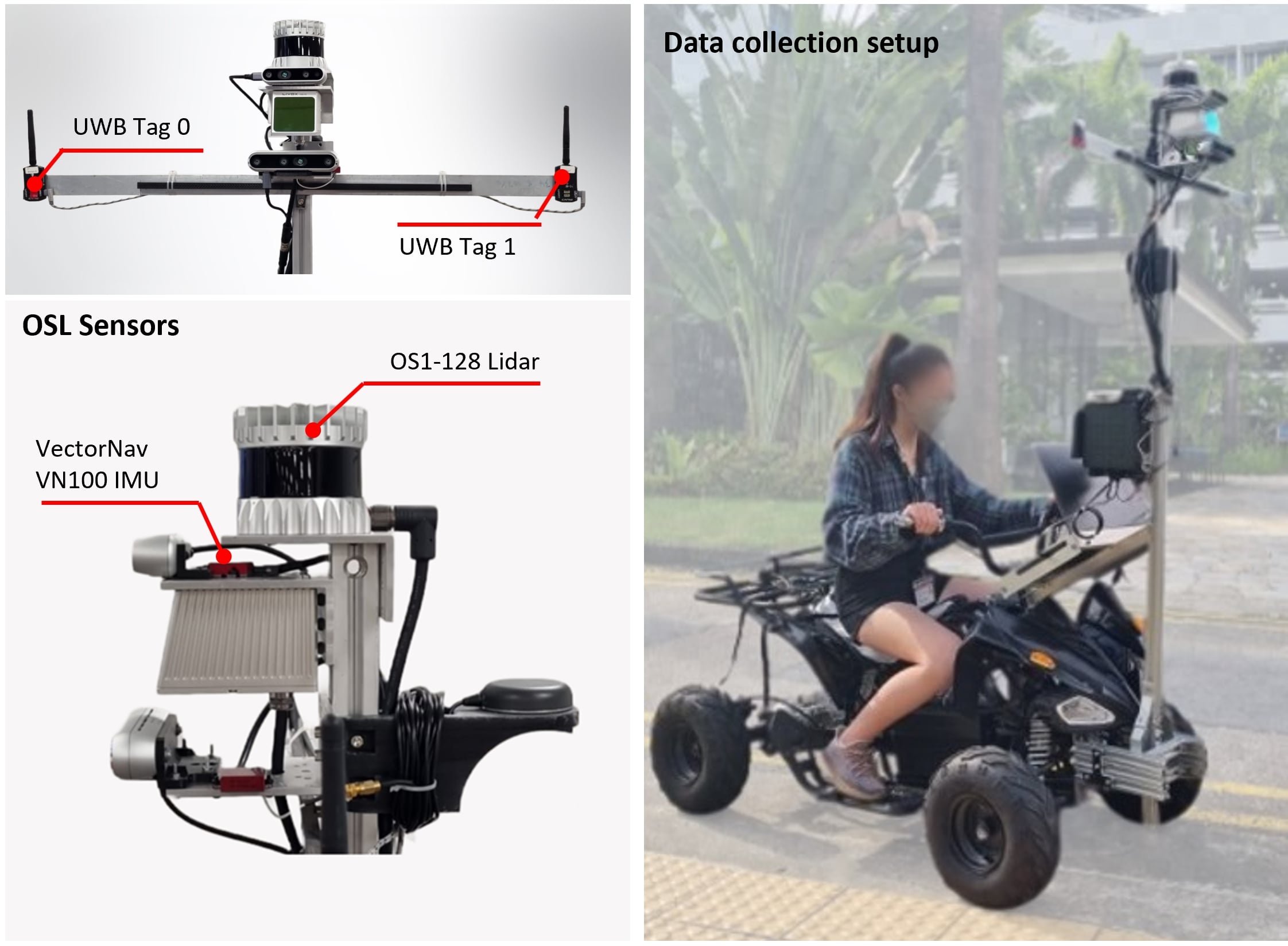}
    \caption{The data collection setup.}
    \label{fig: setup}
\end{figure}

\begin{figure}
    \centering
    \includegraphics[width=\linewidth]{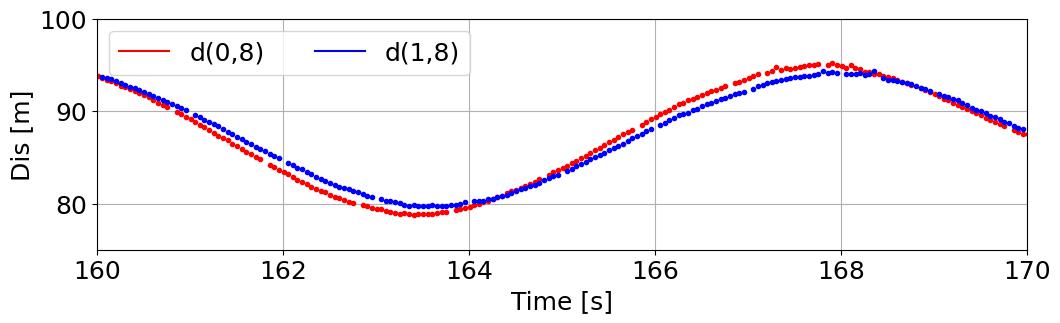}
    \caption{The subtle difference between range measurements by two UWB tags can contain information on the orientation $\rot$ as explained in Fig. \ref{fig: small scale environment}.}
    \label{fig: distance two tags}
\end{figure}

Fig. \ref{fig: setup} illustrates the setup for our data collection scenario. It consists of one LiDAR and one IMU used for the OSL. In addition, two UWB nodes named Tag 0 and Tag 1 are installed on the setup to couple the vehicle's orientation with the range measurements, as explained in Fig. \ref{fig: distance two tags}.

\subsection{Prior map-based localization}

\def\frL{\texttt{L}}
\def\frG{\texttt{G}}

\begin{figure}
    \centering
    \begin{overpic}[width=\linewidth,
                   ]{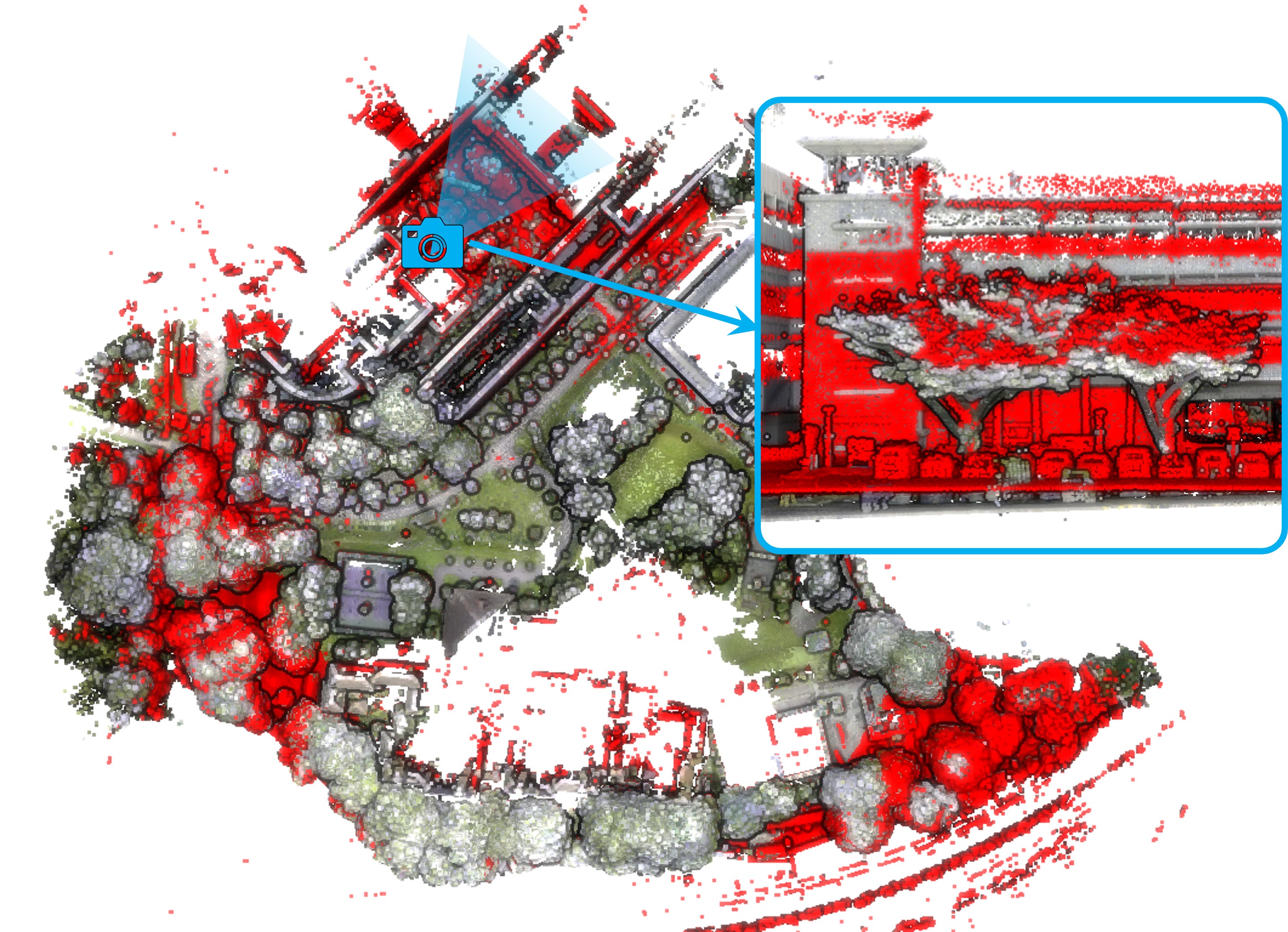}
                   \put(95, 10){(a)}
    \end{overpic}
    \begin{overpic}[width=\linewidth,
                   ]{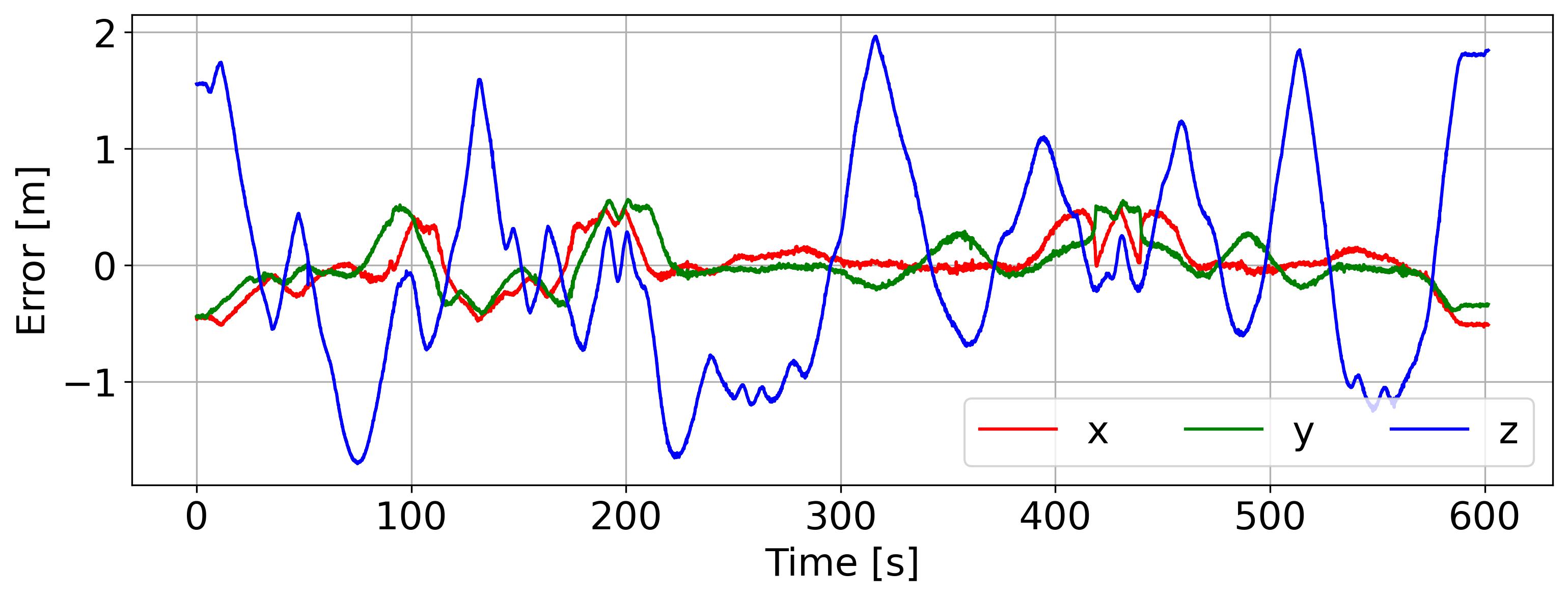}
                   \put(95, 70){(c)}
    \end{overpic}
    \caption{Top-left, a survey-grade prior map (SGPM, in RGB) compared with a loop-closure-enabled prior map from SLICT \cite{nguyen2023slict} (in red). We find that the SLICT-built prior map deviates from the SGPM at certain locations, especially in the z-direction, as shown in the bottom-left image.}
    \label{fig: error distribution}
\end{figure}

In Fig. \ref{fig: error distribution}, we enable the loop-closure feature of SLICT \cite{nguyen2023slict} to build a prior map on one trial of the MCD dataset \cite{mcdviral2024}. Note that this prior map is built entirely based on the OSL data, specifically the point clouds from the Ouster LiDAR and the onboard IMU. SLAM takes the initial location as the origin of its frame of reference. The prior map essentially creates a frame of reference $\frL$. After running the trial, we find the optimal transform ${}^\frL_\frG\tf$ between the prior map coordinate frame $\frL$ and the coordinate frame $\frG$ of ground truth, i.e., the survey-grade prior map. As can be seen in Fig. \ref{fig: error distribution} the error between the prior map $\frL$ and the ground truth map $\frG$ is mainly in the z-direction, most clearly seen where the red points cover up the $\frG$ map. On the other hand, the xy error is well-constrained within 0.5m.

\begin{figure*}
    \centering
    \includegraphics[width=0.49\linewidth]{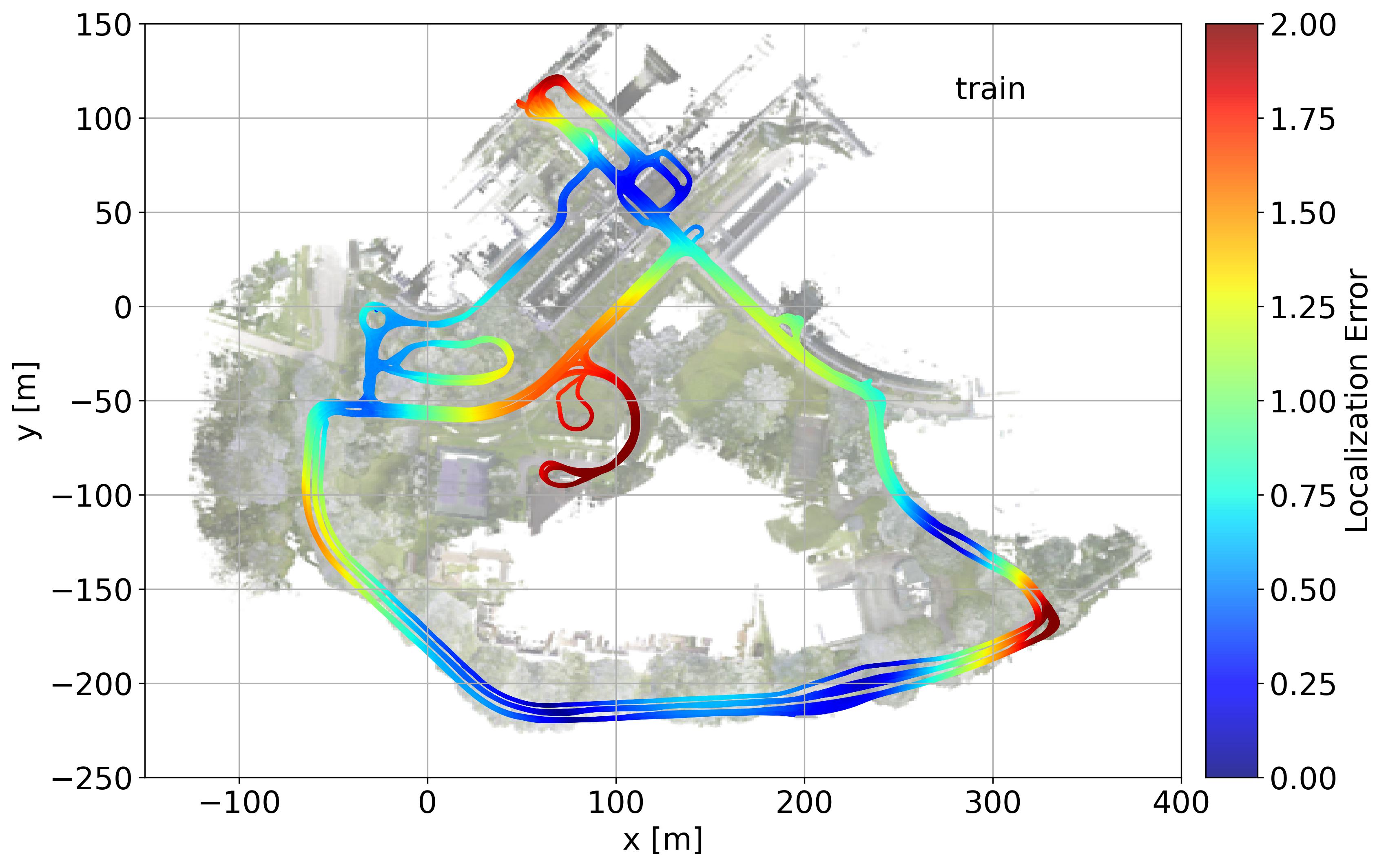}
    \includegraphics[width=0.49\linewidth]{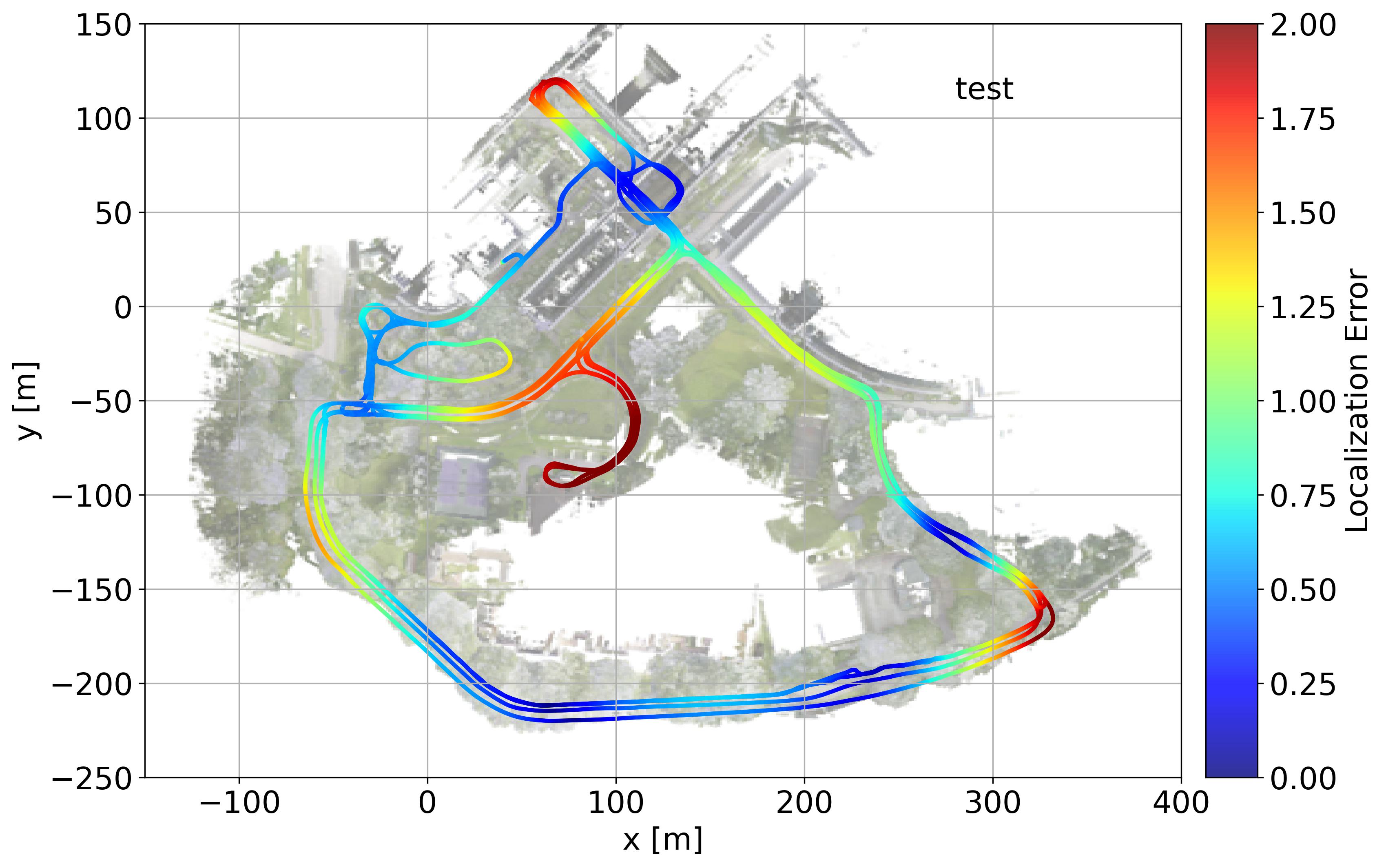}
    \caption{The training trials (top) and the test trials (bottom).}
    \label{fig: train vs test}
\end{figure*}

Next, we run 23 trials with the prior map to obtain the OSL pose estimates and calculate their error. We manually set the initial location of the OSL setup using the ground truth pose.
Out of 23 sequences, 18 are chosen for training and 5 for testing.
As shown in Fig. \ref{fig: train vs test}, there is a strong correlation between the localization error and the actual location, as the same level of error can be seen across the trials at the same location.

Out of 23 trials, data from 18 sequences will be used for training, and 5 will be used for testing. Fig. \ref{fig: train vs test} presents the paths traveled by the vehicle in the training and testing trials along with the localization error.

\subsection{The learning model}

\begin{figure*}
    \centering
    \includegraphics[width=\linewidth]{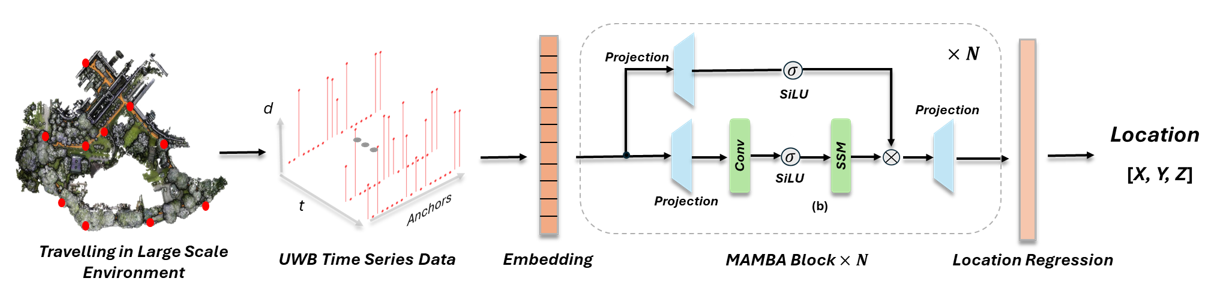}
    \caption{The design of ULOC learning model.}
    \label{fig: learning model}
\end{figure*}

Fig. \ref{fig: learning model} presents an overview of our learning model.
For each trial, we combine all UWB measurements in each 50-ms interval into a vector $X_i = (d_0^0, \dots d_{9}^0, d_0^1, \dots d_{10}^1) \in \R^{20}$, which contains 20 distance measurements from the 2 UWB tags to the ten anchors. If there is no range to an anchor in this interval, the corresponding distance measurement will be set to zero.

For each observation $X_i$, we can obtain a pose estimate $Y_i = (\pos_i^0, \pos^1_i)$, where $\pos^0_i, \pos^1_i \in \R^3$ are the positions of Tag 0 and Tag 1 in the prior map $\frL$ at time step $i$ (SLICT is chosen for this as it offers continuous-time trajectory that allows us to sample the trajectory at any time instance).

As mentioned earlier, we posit that the change in $X_i$ contains contextual information that can be exploited to infer the location. For example, in a sequence of range measurements, the absence of certain anchors can narrow down the area where the robot actually is. Thus, we define a sequence length $S$ (in the $t$ dimension as illustrated in Fig. \ref{fig: learning model}), then from the samples $\{(X_i, Y_i)\}_{i=1}^K$, we can construct $M = K-S+1$ sequences $\{(\X_i, \Y_i)\}_{i=1}^{M}$, where $\X_i = (X_i, \dots X_{i+S-1})$ and $\Y_i = (Y_i, \dots Y_{i+S-1})$.

To fully exploit the contextual information of the input UWB sequence $\X_i$, we adopt MAMBA\cite{gu2023mamba} as our backbone for feature extraction. The MAMBA block is constructed upon the State Space Model (SSM), which captures the temporal patterns of time series data through latent state transition and their relationship with observational measurements in the following dynamics:
\begin{equation}
    \begin{aligned}
        h'(t) &= \boldsymbol{A} h(t) + \boldsymbol{B} x(t) \\
        y(t) &= \boldsymbol{C} h(t) ,
    \end{aligned}
\end{equation}
where $\boldsymbol{A}$, $\boldsymbol{B}$, and $\boldsymbol{C}$ represent the state transition matrix, input matrix, and output matrix, respectively, while $h(t)$ and $y(t)$ denote the hidden state and output at time $t$. For discretization, the continuous sequence is then converted into a discretized SSM using a step size $\Delta$:

\begin{equation}
    \begin{aligned}
        h_t &= \boldsymbol{\overline{A}} h_{t-1} + \boldsymbol{\overline{B}} x_t \\
        y_t &= \boldsymbol{C} h_t,
    \end{aligned}
\end{equation}
where 
$\overline{\boldsymbol{A}} = \exp(\Delta \boldsymbol{A})$ and $
\overline{\boldsymbol{B}} = (\Delta \boldsymbol{A})^{-1} (\exp(\Delta \boldsymbol{A}) - \boldsymbol{I}) \cdot \Delta \boldsymbol{B}$.
To dynamically select the most critic information in time sequence data, $\Delta$, $\boldsymbol{B}$, $\boldsymbol{C}$ are time-varying variables, which are learned through fully connected layers in MAMBA. This adaptability allows the network to focus on other time step data when certain UWB anchors become invalid upon transitioning into a LOS region and to effectively filter out potential outliers caused by multipath effects.

The UWB sequence $\X_i$ is initially input into an embedding layer that transforms the UWB data from a sparse space to a dense feature space. A learnable time-position embedding is then added to further help the network capture contextual information. The UWB embedding subsequently passes through four MAMBA blocks to learn the contextual dependencies within the UWB sequence. The resulting features are finally fed into a fully connected layer to generate predictions on the sequence of vehicle poses $\hat{\Y}_i$. The learning process will optimize the model parameters to minimize the MSE error between $\hat{\Y}_i$ and $\Y_i$. In other words, the learning process can be stated as:
\begin{equation}
    \argmin_{\theta} \sum_{i=1}^{M} \norm{\mathcal{M}(\theta, \X_i) - \Y_i}^2,
\end{equation}
where $\theta$ represent all parameters of the learning model. We note that our model has fewer parameters than the existing LSTM-based model from the literature \cite{lim2019ronet} (more details in the experiments of Sec. \ref{sec: experiments}).

\begin{figure*}
    \centering
    \includegraphics[width=0.32\linewidth]{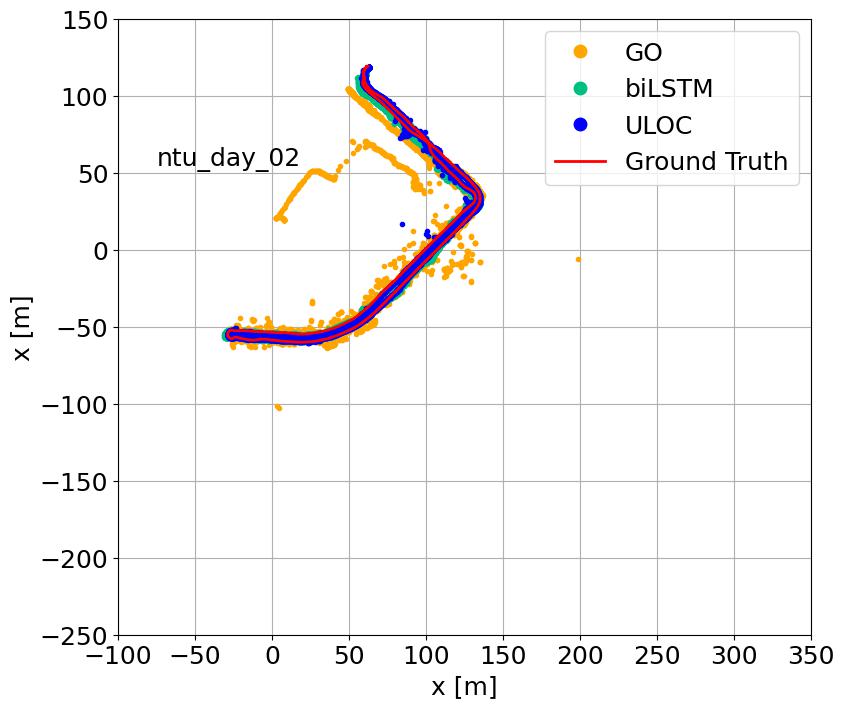}
    \includegraphics[width=0.32\linewidth]{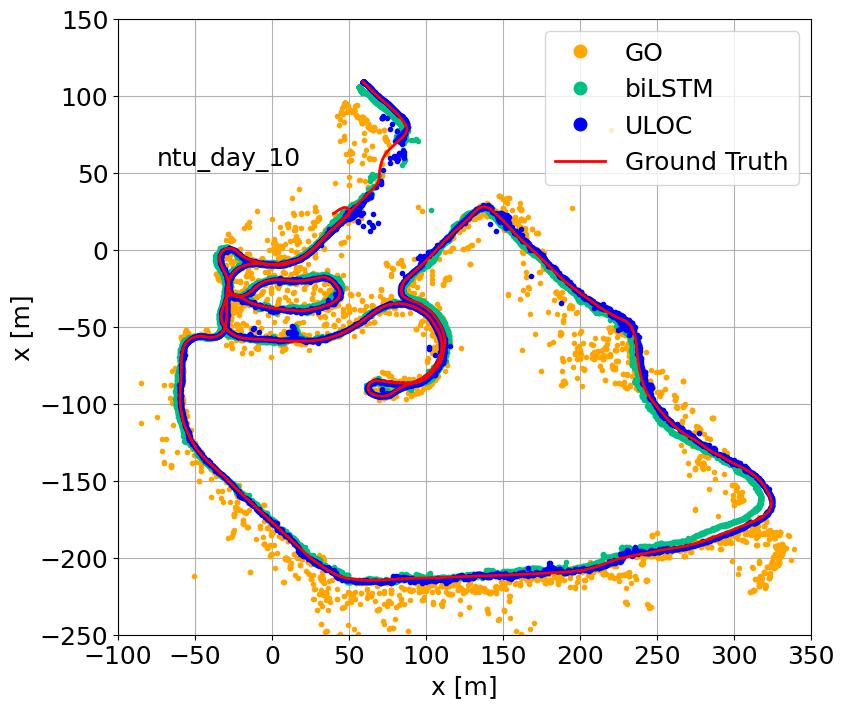}
    \includegraphics[width=0.32\linewidth]{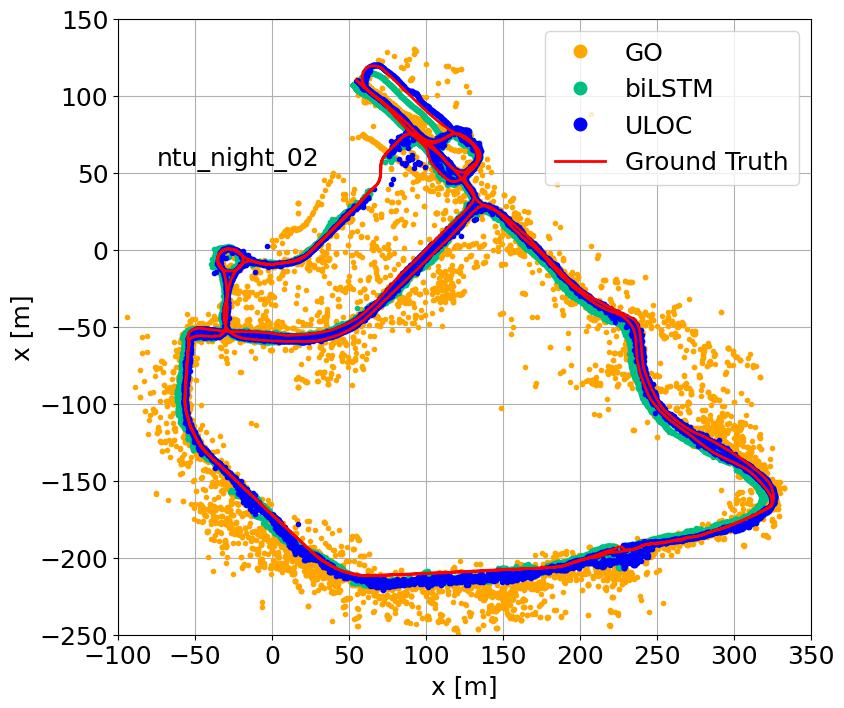}
    
    \includegraphics[width=0.32\linewidth]{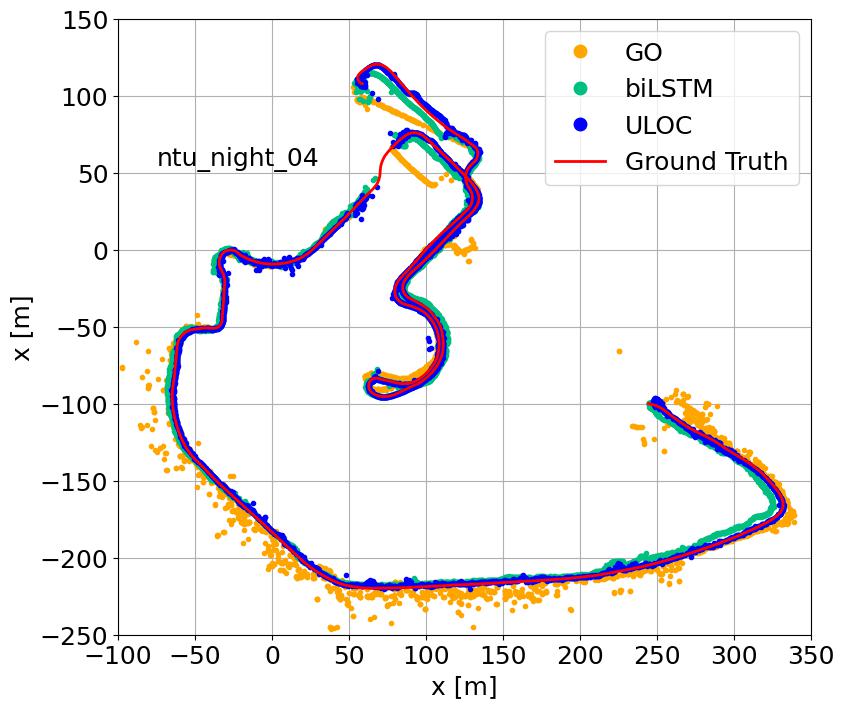}
    \includegraphics[width=0.32\linewidth]{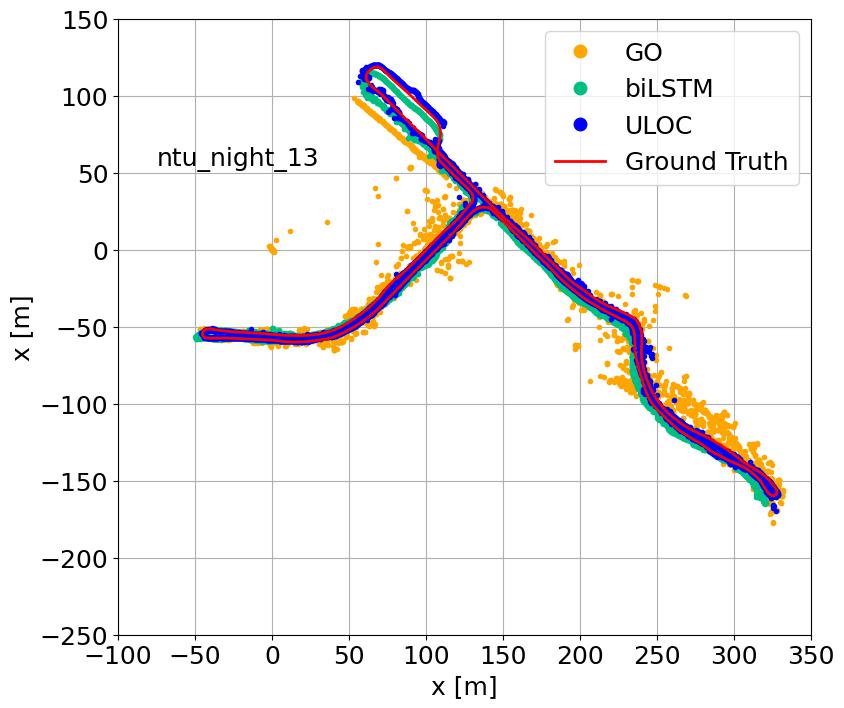}
    \includegraphics[width=0.32\linewidth]{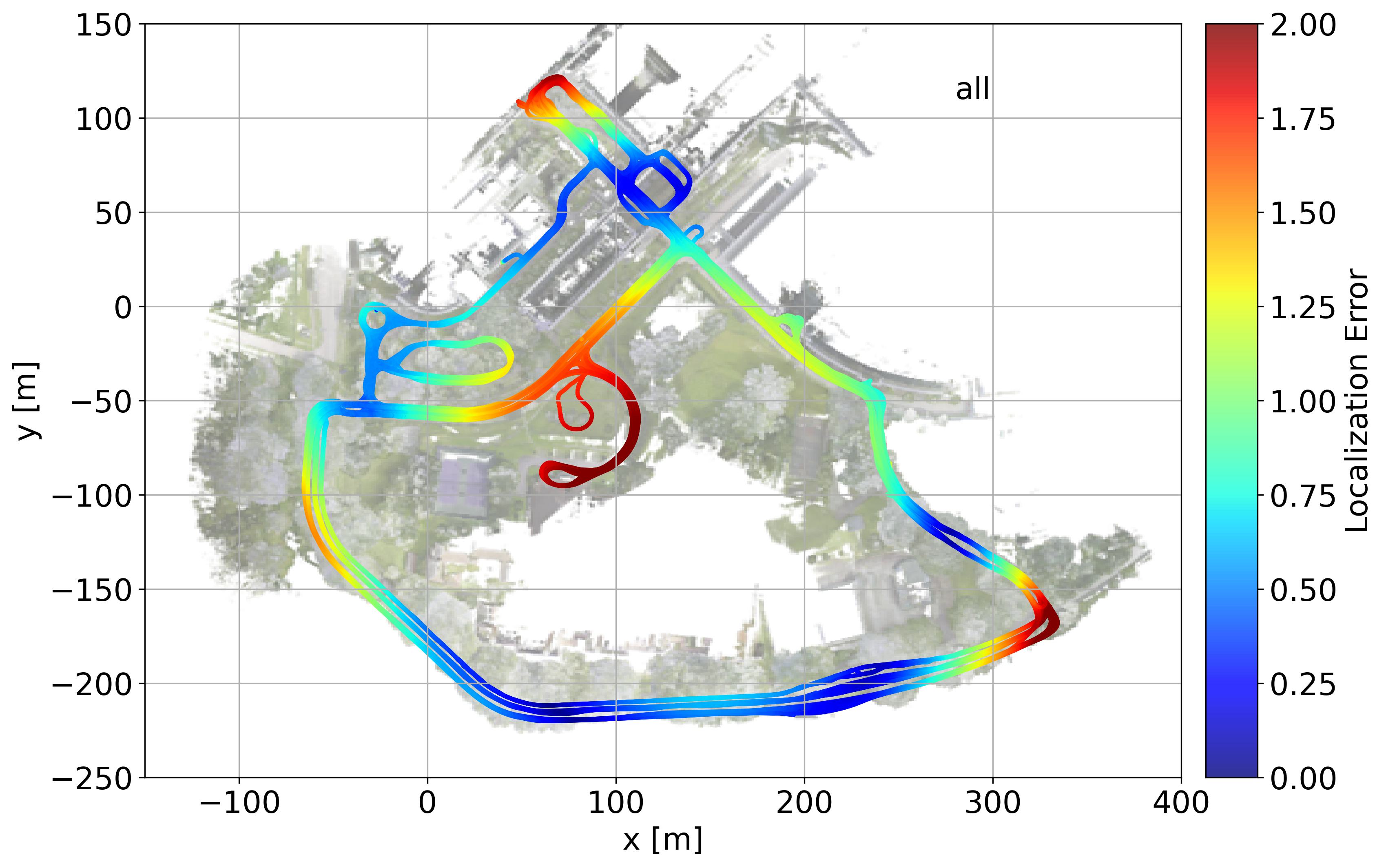}
    \caption{Localization results of different methods. The graph optimization method has very erratic estimate, despite requiring true anchor positions. On the other hand, Both learning methods can follow the ground truth well, however BiLSTM error correlates with the prior map estimation error, while ULOC follows this estimate.}
    \label{fig: 3D error}
\end{figure*}

\section{Experiments} \label{sec: experiments}

\subsection{Experimental setup}

We employ the Adam optimizer with with a batch size of 64 and an initial learning rate of 0.001 for training the models. The models are trained for a total of 150 epochs, with the learning rate reduced by a factor of 0.5 every 20 epochs. To minimize the impact of randomness, each model is trained five times, and the results are averaged. Root Mean Square Error (RMSE) is used as the evaluation metric, which is defined by:
\begin{equation}
\text{RMSE} = \sqrt{\frac{1}{n} \sum_{i=1}^{n} (\hat{Y_i} -{Y_i})^2},
\end{equation}
where $n$ denotes the number of test samples.

\subsection{Comparison with state of the art}

We first conduct an experiment on a small-scale dataset provided by RONET \cite{lim2019ronet}, which consists of measurements to eight anchors in a 6m$\times$6m areas, and only 2D ground-truth is provided. As no information on anchors is provided, we cannot experiment with any classical methods. Rather, we experiment with different learning models, namely GRU, LSTM, BiLSTM, and MAMBA. The result are summarized in Tab. \ref{tab: RONET RMSE}.
It is clear from Tab. \ref{tab: RONET RMSE} that BiLSTM and MAMBA have the best performance among all learning methods. Moreover, we obtain an RMSE of 3.06cm from ULOC (MAMBA), significantly lower than 3.77cm of BiLSTM, and 3.09cm as reported in \cite{lim2019ronet}. This reinforces our choice for MAMBA as new model for UWB-based localization.

Next, we demonstrate the performance of ULOC on large scale dataset. As the experiment with small scale environment has confirmed the avantage of BiLSTM and MAMBA, we only conduct experiment with these two models for the large scale dataset. Moreover, we also include experiment with graph optimization (GO) \cite{wang2017ultra}, a classical method. Tab. \ref{tab: RMSE} reports the RMSE of methods on the test trials. Fig. \ref{fig: 3D error} visualizes the estimates from GO, BiLSTM, ULOC and groundtruth on all of the test trials.

\begin{table}
	\centering
	\begin{threeparttable}
		\caption{RMSE of the test trials with from different methods}
		\label{tab: RONET RMSE}
		
		\begin{tabular}{l|cccc}
			\hline
			\textbf{Test Trials}
			& \textbf{GRU}
			& \textbf{LSTM}
			& \textbf{BiLSTM}
			& \textbf{ULOC} \\
			\hline
			{test\_seq1}
                & 4.11 
                & 4.16
                & \underline{3.82}
                & \textbf{2.85}
                \\
			{test\_seq2}
                & 4.22
                & 4.54
                &
                \underline{3.69}
                & \textbf{3.39}
                \\
			{Average}
                & 4.16
                & 4.30
                & \underline{3.77}
                &
                \textbf{3.06}
                \\
			\hline
		\end{tabular}
		
		\begin{tablenotes}
			\footnotesize
			\item All values are in [cm] unit. The best results are in bold, and second best results are underlined.
		\end{tablenotes}
	\end{threeparttable}
\end{table}

\begin{table}
    \centering
    \begin{threeparttable}
        \caption{RMSE of the test trials with from different methods}
        \label{tab: RMSE}

        \begin{tabular}{l|ccc}
            \hline
              \textbf{Test Trials}
            & \textbf{GO}
            & \textbf{BiLSTM}
            & \textbf{ULOC} \\
            \hline
            {ntu\_day\_02}   & 19.97   &    \underline{3.65}     & \textbf{1.59}            \\
            {ntu\_day\_10}   & 22.55   &    \underline{7.17}     & \textbf{2.90}               \\
            {ntu\_night\_02} & 22.12   &   \underline{4.59}      & \textbf{2.42}               \\
            {ntu\_night\_04} & 10.47   &   \underline{4.16}      & \textbf{2.02}              \\
            {ntu\_night\_13} & 16.84   &   \underline{4.23}      & \textbf{1.75}              \\
            \hline
        \end{tabular}

        \begin{tablenotes}
            \footnotesize
            \item All values are in [m] unit. The best results are in bold, and second best results are underlined.
        \end{tablenotes}
    \end{threeparttable}
\end{table}

\begin{figure*}
    \centering
    \includegraphics[width=0.4\linewidth]{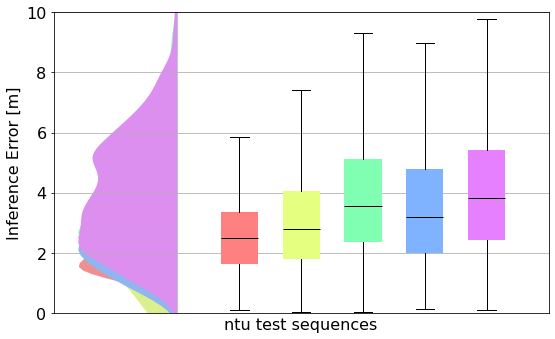}
    \includegraphics[width=0.4\linewidth]{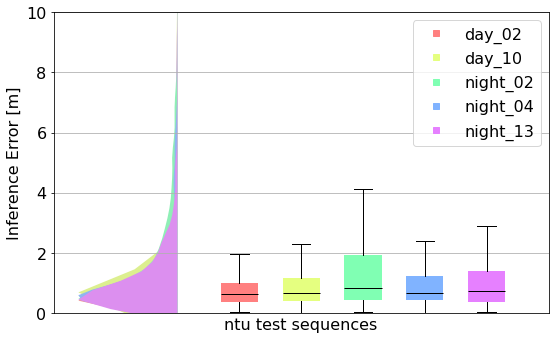}
    
    \caption{The spread of the localization errors by two learning models (left: BiLSTM, right: ULOC). Note that most of the ULOC predictions have an error below 3m.}
    \label{fig: violin}
\end{figure*}

\begin{figure*}
    \centering
    \includegraphics[width=0.4\linewidth]{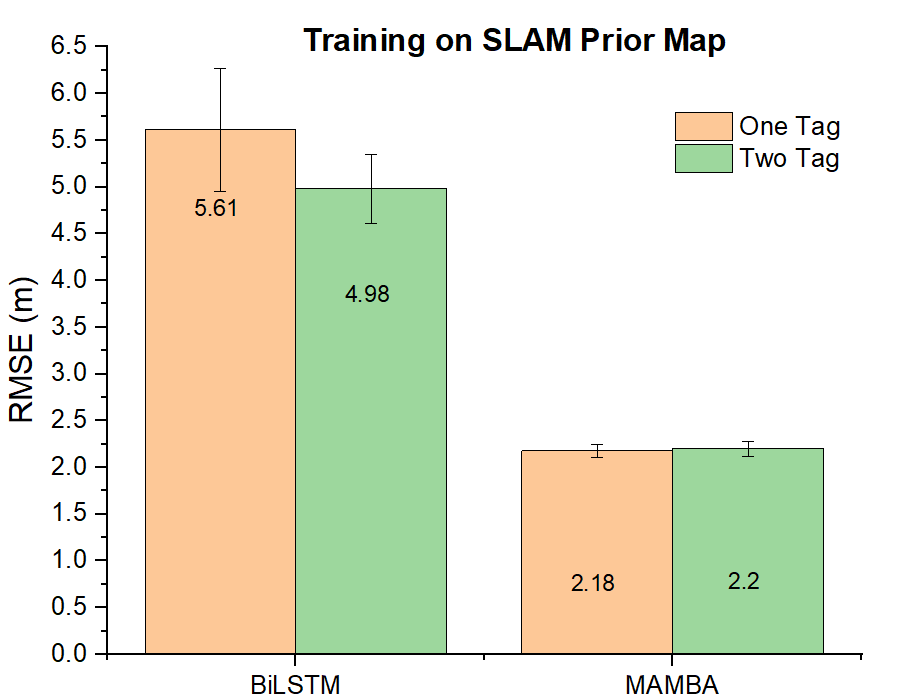}
    \includegraphics[width=0.4\linewidth]{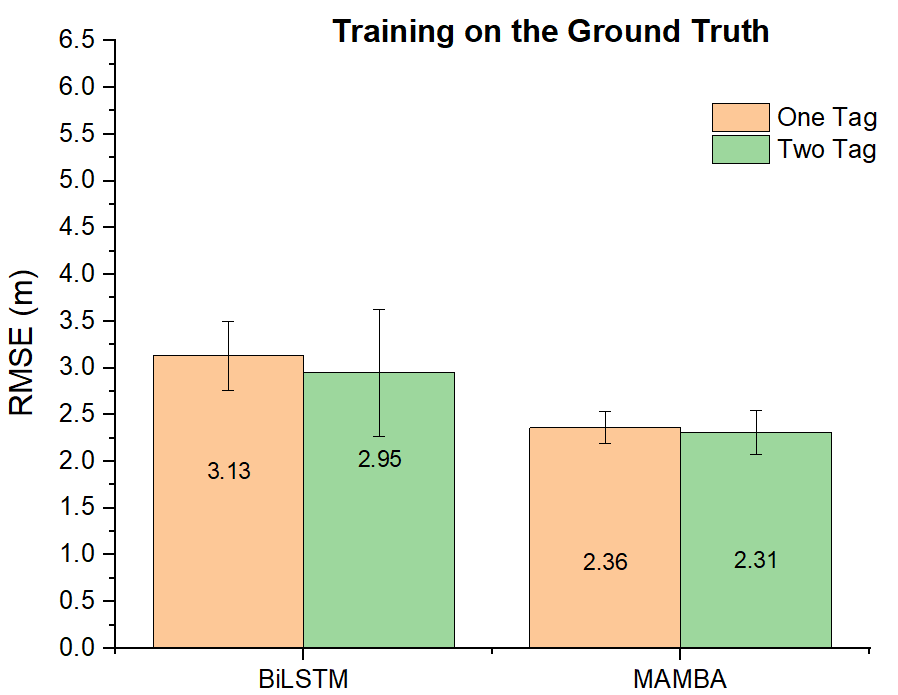}
    
    \caption{RMSE on all test trials with different training data $Y$.}
    \label{fig: ablation study}
\end{figure*}


From Tab. \ref{tab: RMSE}, we can see that even though GO requires anchor location, it has very erratic estimates, due to the simple approach that only focuses on geometry as presented in Fig. \ref{fig: small scale environment}. On the other hand, both learning methods have much better accuracy compared to traditional GO based method. Between BiLSTM and ULOC, we clearly demonstrate the advantage of MAMBA over LSTM approach. Interestingly, from Fig. \ref{fig: 3D error}, we notice that BiLSTM's prediction error has strong correlation with the prior map's error as shown in Fig. \ref{fig: error distribution}. This seems to indicate that BiLSTM is more well-suited for learning the implicit geometry, while MAMBA is more flexible and can better learn the range-location context.
In Fig. \ref{fig: violin}, we present the spread of the localization error of the two learning models across the five test trials.

\subsection{Ablation study}

To confirm the effects of location-dependent bias in $Y$, we retrain the learning models with true ground truth for $Y$ instead of the priorm map based OSL. Moreover, we also conduct another ablation study where only one tag (Fig. \ref{fig: setup}) is used. The results of these two ablation studies are presented in Fig. \ref{fig: ablation study}.

\subsubsection{Effects of prior map based OSL localization.}

From Fig. \ref{fig: ablation study} we can see that the RMSE of BiLSTM reduces significantly from 4.98m to 3.13m when using the ground truth. Interestingly, the RMSE of MAMBA remains almost the same. Given that MAMBA actually has fewer parameters than biLSTM ($\sim$400,000 vs $\sim$600,000), these result indicate that MAMBA is much more effective in learning the context of the time series data, whereas LSTM is compatible with learning the implicit geometry in the problem.

\subsubsection{Effect of single and dual tag}

From. Fig. \ref{fig: ablation study} we find that the use of either one or two tags don't have much influence on the accuracy of MAMBA, but has a more noticeable influence on LSTM. Again, this may be explained if we follow the thesis of LSTM being suited for learning the implicit geometry. On the other hand, as the separation between the two tags is almost equal to the RMSE, this baseline may not contribute much to the prediction accuract by MAMBA. Perhaps when the baseline is several meters large, as when the tags are mounted on a car or truck, their effectiveness can be better observed with MAMBA.

\section{Conclusions and Future Works} \label{sec: conclusion}

In this paper we propose a learning based framework for localization on complex large scale environments using UWB named ULOC. By leveraging OSL with prior map, we obtain localization estimate with location-consistent displacement. We demonstrate that this hidden complexity along with other effects such as NLOS, multi-path, and ranging context can be effectively learned by a RNN model leveraging SOTA architecture MAMBA. Our experiments show that ULOC outperforms SOTA of both classical and learning methods. Ablation study also confirms that MAMBA can effectively learn the ranging context effectively. We release our source code and data to benefit the community.

For the future works, we look forward to realizing a confidence metric of the UWB estimate. This information is crucial in determining whether UWB-based prediction can be trusted for relocalization on the large-scale environments. We also look forward to realizing a long-term prior map OSL system based on SLICT, leveraging ULOC as a the relocalization trigger.

\balance
\bibliographystyle{IEEEtran}
\bibliography{references,extrabib}

\begin{thebibliography}{10}
\providecommand{\url}[1]{#1}
\csname url@samestyle\endcsname
\providecommand{\newblock}{\relax}
\providecommand{\bibinfo}[2]{#2}
\providecommand{\BIBentrySTDinterwordspacing}{\spaceskip=0pt\relax}
\providecommand{\BIBentryALTinterwordstretchfactor}{4}
\providecommand{\BIBentryALTinterwordspacing}{\spaceskip=\fontdimen2\font plus
\BIBentryALTinterwordstretchfactor\fontdimen3\font minus \fontdimen4\font\relax}
\providecommand{\BIBforeignlanguage}[2]{{%
\expandafter\ifx\csname l@#1\endcsname\relax
\typeout{** WARNING: IEEEtran.bst: No hyphenation pattern has been}%
\typeout{** loaded for the language `#1'. Using the pattern for}%
\typeout{** the default language instead.}%
\else
\language=\csname l@#1\endcsname
\fi
#2}}
\providecommand{\BIBdecl}{\relax}
\BIBdecl

\bibitem{nguyen2023vr}
T.~H. Nguyen, S.~Yuan, and L.~Xie, ``Vr-slam: A visual-range simultaneous localization and mapping system using monocular camera and ultra-wideband sensors,'' \emph{arXiv preprint arXiv:2303.10903}, 2023.

\bibitem{jiang2023efficient}
H.~Jiang, W.~Wang, Y.~Shen, X.~Li, X.~Ren, B.~Mu, and J.~Wu, ``Efficient planar pose estimation via uwb measurements,'' in \emph{2023 IEEE International Conference on Robotics and Automation (ICRA)}.\hskip 1em plus 0.5em minus 0.4em\relax IEEE, 2023, pp. 1954--1960.

\bibitem{xu2022omni}
H.~Xu, Y.~Zhang, B.~Zhou, L.~Wang, X.~Yao, G.~Meng, and S.~Shen, ``Omni-swarm: A decentralized omnidirectional visual--inertial--uwb state estimation system for aerial swarms,'' \emph{Ieee transactions on robotics}, vol.~38, no.~6, pp. 3374--3394, 2022.

\bibitem{zhou2022swarm}
X.~Zhou, X.~Wen, Z.~Wang, Y.~Gao, H.~Li, Q.~Wang, T.~Yang, H.~Lu, Y.~Cao, C.~Xu \emph{et~al.}, ``Swarm of micro flying robots in the wild,'' \emph{Science Robotics}, vol.~7, no.~66, p. eabm5954, 2022.

\bibitem{moron2022towards}
P.~T. Mor{\'o}n, J.~P. Queralta, and T.~Westerlund, ``Towards large-scale relative localization in multi-robot systems with dynamic uwb role allocation,'' in \emph{2022 7th International Conference on Robotics and Automation Engineering (ICRAE)}.\hskip 1em plus 0.5em minus 0.4em\relax IEEE, 2022, pp. 239--246.

\bibitem{wang2023stop}
X.~Wang, C.~Jiang, S.~Sheng, and Y.~Xu, ``Stop-line-aided cooperative positioning for connected vehicles,'' \emph{IEEE Transactions on Intelligent Vehicles}, vol.~8, no.~2, pp. 1765--1776, 2023.

\bibitem{nguyen2016ultra}
T.-M. Nguyen, A.~H. Zaini, K.~Guo, and L.~Xie, ``An ultra-wideband-based multi-uav localization system in gps-denied environments,'' in \emph{International Micro Air Vehicle Competition and Conference 2016}, Beijing, China, Oct. 2016, pp. 56--61.

\bibitem{nguyen2021ntuviral}
T.-M. Nguyen, S.~Yuan, M.~Cao, Y.~Lyu, T.~H. Nguyen, and L.~Xie, ``Ntu viral: A visual-inertial-ranging-lidar dataset, from an aerial vehicle viewpoint,'' \emph{The International Journal of Robotics Research}, vol.~41, no.~3, pp. 270--280, 2022.

\bibitem{zhao2022util}
W.~Zhao, A.~Goudar, X.~Qiao, and A.~P. Schoellig, ``Util: An ultra-wideband time-difference-of-arrival indoor localization dataset,'' \emph{The International Journal of Robotics Research}, p. 02783649241230640, 2022.

\bibitem{yuan2023std}
C.~Yuan, J.~Lin, Z.~Zou, X.~Hong, and F.~Zhang, ``Std: Stable triangle descriptor for 3d place recognition,'' in \emph{2023 IEEE international conference on robotics and automation (ICRA)}.\hskip 1em plus 0.5em minus 0.4em\relax IEEE, 2023, pp. 1897--1903.

\bibitem{yin2024outram}
P.~Yin, H.~Cao, T.-M. Nguyen, S.~Yuan, S.~Zhang, K.~Liu, and L.~Xie, ``Outram: One-shot global localization via triangulated scene graph and global outlier pruning,'' in \emph{2024 IEEE International Conference on Robotics and Automation (ICRA)}.\hskip 1em plus 0.5em minus 0.4em\relax IEEE, 2024, pp. 13\,717--13\,723.

\bibitem{gu2023mamba}
A.~Gu and T.~Dao, ``Mamba: Linear-time sequence modeling with selective state spaces,'' \emph{arXiv preprint arXiv:2312.00752}, 2023.

\bibitem{mueller2015fusing}
M.~W. Mueller, M.~Hamer, and R.~D'Andrea, ``Fusing ultra-wideband range measurements with accelerometers and rate gyroscopes for quadrocopter state estimation,'' in \emph{2015 IEEE International Conference on Robotics and Automation (ICRA)}.\hskip 1em plus 0.5em minus 0.4em\relax IEEE, 2015, pp. 1730--1736.

\bibitem{tiemann2017scalable}
J.~Tiemann and C.~Wietfeld, ``Scalable and precise multi-uav indoor navigation using tdoa-based uwb localization,'' in \emph{2017 International Conference on Indoor Positioning and Indoor Navigation (IPIN)}.\hskip 1em plus 0.5em minus 0.4em\relax IEEE, 2017, pp. 1--7.

\bibitem{dai2024cubature}
C.~He, C.~Tang, and C.~Yu, ``A federated derivative cubature kalman filter for imu-uwb indoor positioning,'' \emph{Sensors}, vol.~20, no.~12, p. 3514, 2020.

\bibitem{wang2017ultra}
C.~Wang, H.~Zhang, T.-M. Nguyen, and L.~Xie, ``Ultra-wideband aided fast localization and mapping system,'' in \emph{2017 IEEE/RSJ International Conference on Intelligent Robots and Systems (IROS)}.\hskip 1em plus 0.5em minus 0.4em\relax IEEE, 2017, pp. 1602--1609.

\bibitem{fang2019graph}
X.~Fang, C.~Wang, T.-M. Nguyen, and L.~Xie, ``Graph optimization approach to range-based localization, early access,'' \emph{IEEE Transactions on Systems, Man, and Cybernetics: Systems}, 2020.

\bibitem{zhou2021graph}
Y.~Wang and X.~Li, ``Graph-optimization-based zupt/uwb fusion algorithm,'' \emph{ISPRS International Journal of Geo-Information}, vol.~7, no.~1, p.~18, 2018.

\bibitem{lim2019ronet}
H.~Lim, C.~Park, and H.~Myung, ``Ronet: Real-time range-only indoor localization via stacked bidirectional lstm with residual attention,'' in \emph{2019 IEEE/RSJ International Conference on Intelligent Robots and Systems (IROS)}.\hskip 1em plus 0.5em minus 0.4em\relax IEEE, 2019, pp. 3241--3247.

\bibitem{nosrati2022improving}
L.~Nosrati, M.~S. Fazel, and M.~Ghavami, ``Improving indoor localization using mobile uwb sensor and deep neural networks,'' \emph{IEEE Access}, vol.~10, pp. 20\,420--20\,431, 2022.

\bibitem{zhao2022finding}
W.~Zhao, A.~Goudar, and A.~P. Schoellig, ``Finding the right place: Sensor placement for uwb time difference of arrival localization in cluttered indoor environments,'' \emph{IEEE Robotics and Automation Letters}, vol.~7, no.~3, pp. 6075--6082, 2022.

\bibitem{hashim2023uwb}
H.~A. Hashim, A.~E. Eltoukhy, and K.~G. Vamvoudakis, ``Uwb ranging and imu data fusion: Overview and nonlinear stochastic filter for inertial navigation,'' \emph{IEEE Transactions on Intelligent Transportation Systems}, 2023.

\bibitem{hashim2023nonlinear}
H.~A. Hashim, A.~E. Eltoukhy, K.~G. Vamvoudakis, and M.~I. Abouheaf, ``Nonlinear deterministic observer for inertial navigation using ultra-wideband and imu sensor fusion,'' in \emph{2023 IEEE/RSJ International Conference on Intelligent Robots and Systems (IROS)}.\hskip 1em plus 0.5em minus 0.4em\relax IEEE, 2023, pp. 3085--3090.

\bibitem{li2023continuous}
K.~Li, Z.~Cao, and U.~D. Hanebeck, ``Continuous-time ultra-wideband-inertial fusion,'' \emph{IEEE Robotics and Automation Letters}, vol.~8, no.~7, pp. 4338--4345, 2023.

\bibitem{goudar2023continuous}
A.~Goudar, T.~D. Barfoot, and A.~P. Schoellig, ``Continuous-time range-only pose estimation,'' in \emph{2023 20th Conference on Robots and Vision (CRV)}.\hskip 1em plus 0.5em minus 0.4em\relax Los Alamitos, CA, USA: IEEE Computer Society, June 2023, pp. 29--36.

\bibitem{nguyen2023slict}
T.-M. Nguyen, D.~Duberg, P.~Jensfelt, S.~Yuan, and L.~Xie, ``Slict: Multi-input multi-scale surfel-based lidar-inertial continuous-time odometry and mapping,'' \emph{IEEE Robotics and Automation Letters}, vol.~8, no.~4, pp. 2102--2109, 2023.

\bibitem{mcdviral2024}
\BIBentryALTinterwordspacing
T.-M. Nguyen, S.~Yuan, T.~H. Nguyen, P.~Yin, H.~Cao, L.~Xie, M.~Wozniak, P.~Jensfelt, M.~Thiel, J.~Ziegenbein, and N.~Blunder, ``Mcd: Diverse large-scale multi-campus dataset for robot perception,'' in \emph{Proceedings of the IEEE/CVF Conference on Computer Vision and Pattern Recognition}, Jun. 2024. [Online]. Available: \url{https://mcdviral.github.io/}
\BIBentrySTDinterwordspacing

\end{thebibliography}

\end{document}